\documentclass[accepted]{uai2025} 
                        
\usepackage[american]{babel}

\usepackage{natbib} 
\usepackage{mathtools} 
\usepackage{booktabs} 
\usepackage{tikz} 

\usepackage{amsmath}
\usepackage{amssymb}
\usepackage{mathtools}
\usepackage{amsthm}

\usepackage{url}            
\usepackage{nicefrac}       
\usepackage{microtype}      
\usepackage{xcolor}         
\usepackage{amsmath}
\usepackage{graphicx}
\usepackage{caption}
\usepackage{siunitx}
\usepackage{listings,newtxtt}
\usepackage{algorithmic}
\usepackage{algorithm}
\usetikzlibrary{bayesnet}
\usetikzlibrary{arrows.meta, positioning, shapes, fit, backgrounds, calc, patterns}

\usepackage{subfig}
\usepackage{hyperref}
\definecolor{deepblue}{rgb}{0,0,0.6}
\definecolor{deepred}{rgb}{0.6,0,0}
\definecolor{deepgreen}{rgb}{0,0.5,0}

\lstdefinestyle{lststyle}{
language=Python,
basicstyle=\ttfamily\small,
commentstyle=\color{deepred},
otherkeywords={self},             
keywordstyle=\color{deepgreen},
emphstyle=\color{deepblue},    
frame=tb,                         
showstringspaces=false            %
}

\title{Simulation-based Inference for High-dimensional Data using Surjective Sequential Neural Likelihood Estimation}

\author[1,2]{\href{mailto:<simon.dirmeier@sdsc.ethz.ch>?Subject=Your UAI 2025 paper}{Simon~Dirmeier}{}}
\author[3]{Carlo~Albert}
\author[2,4]{Fernando~Perez-Cruz}

\affil[1]{%
    Swiss Data Science Center, Switzerland
}
\affil[2]{%
    ETH Zurich, Switzerland
}
\affil[3]{%
    Swiss Federal Institute of Aquatic Science and Technology, Switzerland
}
\affil[4]{%
    Bank for International Settlements, Switzerland
}
  
\begin{document}
\maketitle

\begin{abstract}
Neural likelihood estimation methods for simulation-based inference can suffer from performance degradation when the modeled data is very high-dimensional or lies along a lower-dimensional manifold, which is due to the inability of the density estimator to accurately estimate a density function. We present Surjective Sequential Neural Likelihood (SSNL) estimation, a novel member in the family of methods for simulation-based inference (SBI). SSNL fits a dimensionality-reducing surjective normalizing flow model and uses it as a surrogate likelihood function, which allows for computational inference via Markov chain Monte Carlo or variational Bayes methods. Among other benefits, SSNL avoids the requirement to manually craft summary statistics for inference of high-dimensional data sets, since the lower-dimensional representation is computed simultaneously with learning the likelihood and without additional computational overhead. We evaluate SSNL on a wide variety of experiments, including two challenging real-world examples from the astrophysics and neuroscience literatures, and show that it either outperforms or is on par with state-of-the-art methods, making it an excellent off-the-shelf estimator for SBI for high-dimensional data sets.
\end{abstract}

\section{{Introduction}}
In the natural sciences, especially in disciplines such as biology and physics, Bayesian inference is becoming increasingly popular due to its ability both to quantify uncertainty in parameter values and to incorporate prior knowledge about quantities of interest. Bayesian statistics infers the posterior distribution $p( \theta | {y}) \propto p(y | \theta) p( \theta)$ of statistical parameters $\theta$ by conditioning a prior distributions $p( \theta)$ on data ${y}$. If the likelihood function $p({y} | \theta)$ is available, i.e., tractable to compute, conventional Bayesian inference using Markov chain Monte Carlo or variational methods can be used for parameter inference \citep{brooks2011handbook,wainwright2008graphical}. However, for many scientific hypotheses, the likelihood is not easy to compute and the experimenter merely has access to a simulator function ${sim}(\theta)$ that can generate synthetic data conditionally on a parameter configuration $\theta$. 

In the latter case, an emergent family of methods collectively called \textit{simulation-based inference} (SBI, \citet{cranmer2020frontier}) has been proposed. 
Traditionally, approximate Bayesian computation (ABC, \citet{sisson2018handbook}), and most successfully sequential Monte Carlo ABC (SMC-ABC; e.g., \citet{beaumont2009adaptive,lenormand2013adaptive}) or simulated annealing ABC (SABC; e.g., \citet{albert2015simulated}), has been used to infer approximate posterior distributions \citep{pritchard1999population,ratmann2007using}. More recently, methods that are based on neural density or density-ratio estimation have found increased application in the natural sciences due to their reduced computational cost and convincing inferential accuracy \citep{brehmer2018constraining,delaunoy2020lightning,gonccalves2020training,hermans2021towards,brehmer2021simulation,dax2021real}. Among these, several branches of methods exist. Likelihood-based methods \citep{papamakarios2019sequential,glockler2022variational} fit a surrogate model for the likelihood function using neural density estimators \citep{papamakarios2021normalizing} which allows to do conventional Bayesian inference and which has been shown to bring significant performance advantages in comparison to ABC methods with the same computational budget. \citet{cranmer2015approximating,durkan2020contrastive,hermans2020likelihood,thomas2022likelihood,delaunoy2022towards,miller2022contrastive} developed similar methods that instead target the likelihood-to-evidence ratio rather than the likelihood, while \citet{papamakarios2016fast,lueckmann2017flexible,greenberg2019automatic,deistler2022truncated,wildberger2023flow,sharrock2024sequential} developed methods that try to approximate the posterior distribution directly.

In the case of likelihood-based methods, the accuracy of posterior inferences might suffer due to the inability of neural density estimators to correctly approximate the surrogate likelihoods, e.g., if the data are very high-dimensional or the data are embedded in a low-dimensional manifold but lie in a higher-dimensional ambient space \citep{fefferman2016testing,kingma2018glow,greenberg2019automatic,cunningham2020normalizing,dai2020sliced,klein2021funnels}. 

To overcome this limitation, we present a new method for simulation-based inference which we call \textit{Surjective Sequential Neural Likelihood} (SSNL) estimation. SSNL uses a surjective dimensionality-reducing normalizing flow to model the surrogate likelihood of a Bayesian model by that allowing improved density estimation and consequently improved posterior inferences. We evaluate SSNL on multiple experiments from the SBI, astrophysics and neuroscience literatures and demonstrate that it achieves superior performance in comparison to state-of-the-art methods. Conversely, we also demonstrate negative examples when our method should, in theory and empirically, not have a performance gain.
\begin{figure*}[h!]
\centering
\subfloat[Bijection.]{
\includegraphics[width=0.7\columnwidth]{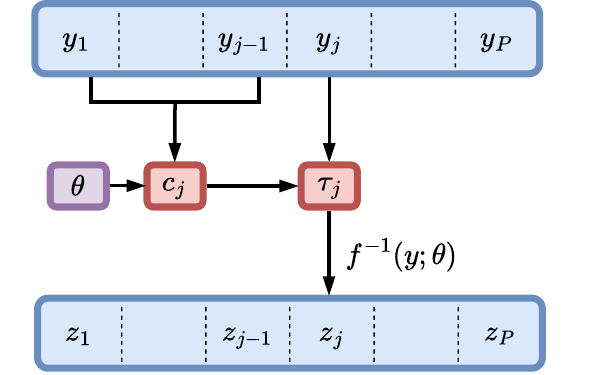}
\label{fig:bijection}%
}
\centering
\subfloat[Surjection.]{%
\includegraphics[width=0.7\columnwidth]{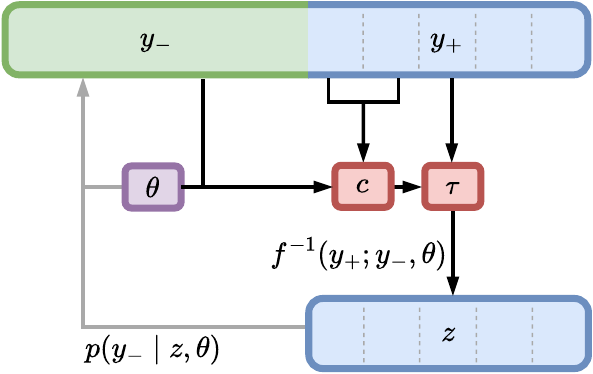}
\label{fig:surjection}%
}
\caption{Conditional bijective and surjective flow layers illustrated with masked autoregressive flows. (a) A bijective flow layer $f^{-1}(y; \theta)$ transforms an input $y_j$ conditional on all previous values $y_{<j}$ and a parameter vector $\theta$ using a conditioner $c$ and transformer $\tau$. (b) The surjective flow layer $f^{-1}(y_{+}; y_{-}, \theta)$ first splits the vector into two components $y_{+}$ and $y_{-}$ and then uses the component $y_{-}$ as additional conditioning variable. The implementations for conditioner and transformer remain the same as for the bijection. To evaluate the likelihood, a surjective layer additionally computes the conditional density $p(y_{-}|z, \theta)$ (which computationally is done after the transform).}
\label{fig:bijection-vs-surjection}
\end{figure*}

\section{{Background}}
\label{sec:background}
Given prior parameter values $\theta \sim p( \theta)$, a simulator function $sim( \theta)$ is a computer program or experimental procedure that can simulate an observation ${y} \leftarrow sim( \theta)$. Apart from stochasticity produced by $p(\theta)$, the simulator might be making use of another source of endogenous randomness. The simulator defines, albeit implicitly, a conditional probability distribution $p({y} | \theta)$ to which the modeller does not have access or which they cannot evaluate in reasonable time. The goal of SBI is to infer the posterior distribution $p( \theta | {y}) \propto p({y} | \theta) p( \theta)$ using synthetic data $\{(y_n, \theta_n) \}_{n=1}^N$ generated from the prior model $p(\theta)$ and simulator $sim(\theta)$. Typically, the total simulation budget $N$ is limited and the posterior for a specific observation ${y}_\text{obs}$ is the target of inference \citep{cranmer2020frontier}. In the following, we introduce relevant background on density estimation with normalizing flows and neural likelihood methods (background on neural posterior and ratio estimation methods can be found in Appendix~\ref{appendix:more-background}).

\subsection{Density estimation using normalizing flows}
Sequential density-based SBI methods (e.g., \citet{greenberg2019automatic,papamakarios2019sequential,deistler2022truncated}) use conditional normalizing flows to fit a surrogate model to either approximate the intractable likelihood or posterior. Normalizing flows (NFs, \citet{papamakarios2021normalizing}) model a probability distribution via a pushforward measure as 
\begin{equation}
q_{f}({y} |  \theta) = p \left( {z}_0 \right) \prod_k^K \big| \det J_k \big|^{-1}
\label{equ:nflikelihood}
\end{equation} 
where $\det J_k = \det \frac{\partial f_k}{\partial {z}_{k-1}}$ is the determinant of the Jacobian matrix of a forward transformation $f_k$ which is typically parameterized with a neural network, and $p \left( {z}_0 \right)$ is some base distribution that has a density that can be evaluated exactly, for instance, a spherical multivariate Gaussian. The forward transformations $f = (f_1, \dots, f_K)$ are a sequence of $K$ diffeomorphisms which are applied consecutively to compute $y = z_K = f_K \circ \dots \circ f_2 \circ f_1(z_0)$. The two densities $q$ and $p$ are related by the multiplicative terms $\det J_k$ which are needed to account for the change-of-volume induced by $f_k$ and which are termed likelihood contribution in \citet{nielsen2020survae} and \citet{klein2021funnels}. The diffeomorphisms $f_k$ are required to be dimensionality-preserving and invertible to be able to both evaluate the probability of a data point and to draw samples. Particularly, in autoregressive flows \citep{kingma2016improved,papamakarios2017masked,germain2015made} each transformation $f_k$ admits a Jacobian determinant which is efficient to compute and which can be decomposed into an autoregressive \textit{conditioner} $c_i$ and an invertible \textit{transformer} $\tau_i$ as
\begin{equation*}
    z_{k,i} = \tau_i(z_{k - 1,i}, c_i(z_{k - 1, < i}))
\end{equation*}
where all $c_i$ can be computed jointly using a masked neural network (Figure~\ref{fig:bijection}).

\subsection{Neural likelihood estimation}
Sequential neural likelihood estimation (SNL, \citet{papamakarios2019sequential}) iteratively fits a density estimator to approximate the likelihood via $q_{f}({y} | \theta) \approx p({y} | \theta)$. SNL proceeds in $R$ rounds distributing the total simulation budget $N$ evenly in each of these:
in the first round, $r=1$, a prior sample $\theta_n \sim p( \theta)$ of size $N_R = N/R$ is drawn and  used to simulate data points $y_n \leftarrow sim(\theta_n)$ yielding the data set $\mathcal{D} = \{ ({y}_n,  \theta_n) \}^r_{1 \dots N_R}$. The simulated data is used to train a conditional normalizing flow  by maximizing the expected probability $\mathbb{E}_{\mathcal{D}} \left[  q_{f}( {y} | \theta) \right]$. Having access to a likelihood approximation, posterior realizations can be generated either by sampling from $\hat{p}^r( \theta | {y}_\text{obs}) \propto q_{f}({y}_\text{obs} | \theta) p( \theta)$ via Markov chain Monte Carlo or via optimization by fitting a variational approximation to the approximate posterior. SNL then uses the surrogate posterior as proposal prior distribution for the next round, $r + 1$, i.e., it draws a new sample of parameters $  \theta_n  \sim \hat{p}^r( \theta | {y}_\text{obs})$ which are then used to simulate a new batch of pairs $\{ ({y}_n,  \theta_n) \}^{r + 1}_{1 \dots N_R}$. The data sets from the previous rounds and the current round are then appended together and a new model is trained on the entire data set. With an infinite simulation budget and a sufficiently flexible density estimator $q_{f}$, SNL converges to the desired posterior distribution $p( \theta | {y}_\text{obs})$. The simulation budget is in practice typically limited, the number simulations should be held as small as possible, and finite data leads to inaccurate approximations to the likelihood functions.

\section{{Surjective Sequential Neural Likelihood Estimation}}
\label{sec:methods}
\textit{Surjective Sequential Neural Likelihood} (SSNL) estimation approximates the intractable likelihood $p(y|\theta)$ function of a Bayesian model $p(\theta|y)$ while simultaneously embedding the data in a lower-dimensional space using dimensionality-reducing surjective flows. We assume that if the data lie in a high-dimensional ambient space, which is the case for many real-world data sets like time series data, embedding them in a lower-dimensional space should improve likelihood estimation and consequently posterior inference.

We motivate the derivation of the surjective flow layer using the holistic generative framework of \citet{nielsen2020survae} which models the log-probability of a $P$-dimensional data point ${y}$ as
\begin{equation*}
\log p({y}) \simeq \log p \left( {z} \right) + V({y}, {z}) + E({y}, {z}) , \quad {z} \sim q({z} | {y})
\end{equation*}
where $q({z} | {y})$ is some amortized (variational) distribution, ${z}$ is a latent variable with distribution $p({z})$,
$V({y}, {z}) = \log \tfrac{p(y|z)}{q(z|y)}$ is denoted \textit{likelihood contribution term} and $E({y}, {z}) = \log \tfrac{q(z|y)}{p(z|y)}$ is a \textit{bound looseness term}. Intringuingly, for inference surjections, i.e., the kind of flow layers we are considering here, the likelihood contribution can be calculated as
\begin{equation*}
V({y}, {z}) = \lim\limits_{q({z} | {y}) \rightarrow \delta \left( {z}  - h^{-1}({y}) \right)} \mathbb{E}_{q({z} | {y})}\left[ \log   \frac{p({y} | {z})}{q({z} | {y})} \right]
\end{equation*}
where $p({y} | {z})$ is a conditional density, $h^{-1}: \mathcal{Y} \rightarrow \mathcal{Z}$ is a dimensionality-reducing mapping, and where we for convenience of notation denote with $h: \mathcal{Z} \rightarrow \mathcal{Y}$ a right inverse function to $h^{-1}$ such that $h^{-1} \circ h = Id_\mathcal{Z}$. Critically, for surjective normalizing flows, the bound looseness equals $E({y}, {z}) = 0$ if a right inverse function $h$ exists.

We design a conditional surjection layer for dimensionality-reduction as follows (Figure~\ref{fig:surjection} for a graphical overview). We first split the
data vector ${y} \in \mathbb{R}^P$ into two subvectors ${y} = \left[ {y}_{-}, {y}_{+} \right]^T$ where $y_+ \in \mathbb{R}^Q$ and $Q$ is a hyperparameter. The subvectors are obtained by (arbitrarily) defining two disjoint permutations $\pi_+ \cup \pi_- = \{1, \dots, P \}$, $\pi_+ \cap \pi_- = \emptyset$, and then setting $y_+ = [y_{\pi_+(1)}, \dots, y_{\pi_+(Q)}]^T$ and $y_- = [y_{\pi_-(1)}, \dots, y_{\pi_-(P - Q)}]^T$. We then construct a conditional normalizing flow $f(z; y_-, \theta)$ (i.e., conditional on $y_-$ and $\theta$) and its inverse $f^{-1}(y_+; y_-, \theta)$ and define
\begin{align*}
    q({z} | {y}) &= \delta \left({z} - f^{-1}(y_+; y_-, \theta) \right) \\
              &= \delta \left({y}_+ - f ( z; y_-, \theta) \right) \big|  \det J^{-1}  \big|^{-1}
\end{align*}
where 
\begin{align*}
    J^{-1} = \frac{\partial f^{-1}(y_+; y_-, \theta) }{\partial {y}_+}    \bigg|_{{y}_+=f({z}; y_-, \theta )}
\end{align*}
is the Jacobian of the inverse mapping (see Appendix~\ref{appendix:surjection-layer} for details). Using this result and the conditional distribution $p({y} | {z}) =p({y}_- | {z}, \theta)$ the likelihood contribution for a surjection layer becomes 
\begin{alignat*}{4}
V({y}, {z}) = &   \lim  \limits_{q({z} | {y}) \rightarrow \delta \left( {z}  - h^{-1}({y}) \right)} \mathbb{E}_{ q({z} | {y})}\left[ \log \frac{p({y} | {z})}{q({z} | {y})} \right] &&\\
        = &  \int \delta \left( {z} - f^{-1}({y}_+ ; y_-, \theta) \right) &&\\ 
          & \quad \log \frac{p({y}_- | {z}, \theta)}{\delta\left( {z} - f^{-1}({y}_+; y_-, \theta) \right)} \mathrm{d}{z} & \\
= &   \log p \left({y}_- | f^{-1}({y}_{+}; y_-, \theta)\right) - \log \big|  \det J^{-1}  \big|^{-1}&&
\end{alignat*}
where we used the change of variables $\tilde{y}^+ = f(z; y_-, \theta)$ yielding $\mathrm{d}\tilde{y}^+ = \mathrm{d}z |\det J^{-1}|^{-1}$. The likelihood of an observation using a surjective flow is consequently the product of three terms:
\begin{equation*}
p \left( {z} \right) p({y}_{-} | {z}, \theta)\big| \det J \big|^{-1}
\end{equation*}
where $p(z)$ is a base distribution, ${z}= f^{-1}( {y}_{+};{y}_{-}, \theta)$ and $\det J = \det \frac{\partial f(\cdot; {y}_{-}, \theta)}{\partial {z}}$ is again the Jacobian determinant of the forward transformation acting on the lower-dimensional vector ${z}$ (see Appendix~\ref{appendix:surjection-layer} for a detailed derivation of the surjection layer). Note that this representation strictly extends the one by \citet{klein2021funnels}, since here we construct flows that are conditioned on the parameter vector $\theta$. Analogously to multi-layered bijective flows (Equation~\eqref{equ:nflikelihood}), the conditional density of a normalizing flow that consists of $K$ dimensionality-reducing layers has the following form:
\begin{equation*}
q_{f}({y} | \theta) = p \left( {z}_0 \right) \prod_k^K p({z}_{k, -} | f_k^{-1}({z}_{k, +};{z}_{k, -}, \theta))
\big| \det J_k \big|^{-1}
\end{equation*}
where $z_{k, -}$ and $z_{k, +}$ are subvectors of $z_k$ that have been constructed as above and $J_k = \frac{\partial f_k(\cdot; {z}_{k, -}, \theta )}{\partial {z}_{k - 1,+}}$ is the Jacobian of the $k$th surjective transformation $f_k$.

For simulation-based-inference, we model the likelihood estimator $q_{f}({y} | \theta)$ as a composition of dimensionality-preserving and -reducing layers:
\begin{equation}
\begin{split}
q_{f}({y} | \theta) = & \ p \left( {z}_0 \right) \prod_{k \in \mathcal{K}_\text{pres}} \big| \det J_k\big|^{-1} \\
& \ \prod_{k \in \mathcal{K}_\text{red}} p({z}_{k, -} | f_k^{-1}({z}_{k, +};{z}_{k, -}, \theta))
\big| \det J_k \big|^{-1}
\end{split}
\label{eqn:ssnl-likelihood}
\end{equation}

where $\mathcal{K}_\text{pres}$ and $\mathcal{K}_\text{red}$ represent sets of indexes for dimensionality-preserving and -reducing flow layers, respectively. For instance, for a total of $K=5$ normalizing flow layers, one could alternate between bijections and surjections by setting the sets $\mathcal{K}_\text{pres} =\{1, 3, 5\}$ and $\mathcal{K}_\text{red} =\{2, 4\}$. Here, we parameterize $f$ using masked autoregressive flows but in general any flow architecture, such as coupling flows \citep{dinh2014nice,dinh2016density}, neural spline flows \citep{durkan2019neural} or neural autoregressive flows \citep{huang2018neural,cao2020block}, is possible.

The dimensionality-reducing flow is fully deterministic in the pullback direction, i.e., in the case of likelihood estimation, but requires sampling from the conditional $p \left( {y}_{-} | z, \theta \right)$ during the forward transformation and, hence, has additional stochastic components other than the base distribution $p({z}_0)$. For our setting, i.e., density estimation, this is however not a limitation.

The lower-dimensional embedding of SSNL solves previous issues of neural likelihood methods when scaled to high-dimensional data sets. In addition, through the dimensionality-reduction the flows require less trainable parameters, which can speed up computation such that more of the computational budget can be used for the simulator or more expressive architectures. The embedding acts, albeit only conceptually, as a collection of summary statistics which consequently replaces the need of manually defining them.

Like other sequential methods, SSNL is trained in $R$ rounds where in every round a new proposal posterior is defined. The proposal posterior can be either sampled from using MCMC methods or approximated with another conditional distribution using variational inference (see Algorithm~\ref{algorithm:ssnl}). 

\begin{algorithm}[tb]
\caption{Surjective sequential neural likelihood}
\label{algorithm:ssnl}
\begin{algorithmic}
   \STATE {\bfseries Inputs:} observation ${y}_\text{obs}$, prior distribution $p( \theta)$, surjective normalizing flow $q_{f}({y} | \theta)$, simulations per round $N_R$, number of rounds $R$
   \STATE {\bfseries Outputs:} approximate posterior distribution $\hat{p}^R( \theta | {y}_\text{obs})$   
   \STATE Initialize proposal $\hat{p}^0( \theta | {y}_\text{obs}) \leftarrow p( \theta)$, data set $\mathcal{D} = \{ \}$
   \FOR{$r \leftarrow 1, \dots, R$}
   \FOR{$n \leftarrow 1, \dots, N_R$}
   \STATE Sample $ \theta_n \sim \hat{p}^{r - 1}( \theta | {y}_\text{obs})$
   \STATE Simulate ${y}_n \leftarrow sim( \theta_n)$ using the simulator function 
   \STATE Concatenate $\mathcal{D} \leftarrow \{\mathcal{D}, ({y}_n, \theta_n) \}$
   \ENDFOR
   \STATE Train $q_{f}({y} |  \theta)$ on $\mathcal{D}$ \\
  \STATE Set $\hat{p}^r( \theta | {y}_\text{obs}) \propto q_{f}({y}_\text{obs} | \theta) p( \theta)$
   \ENDFOR
\end{algorithmic}
\end{algorithm} 

\section{{Experiments}}
\label{sec:experiments}
\begin{figure*}
\begin{center}
\includegraphics[width=\textwidth]{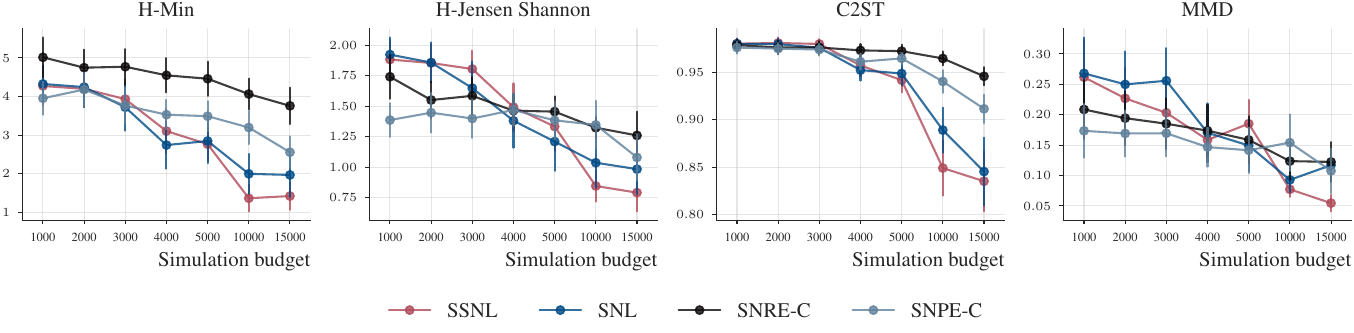}
\caption{Method accuracies on SLCP with different evaluation measures (lower is better; y-axis shows measure).}
\label{fig:slcp_profiles}
\end{center}
\end{figure*}
We compare SSNL to Sequential Neural Likelihood (SNL, \citet{papamakarios2019sequential}), Sequential Neural Posterior Estimation-C (SNPE-C, \citet{greenberg2019automatic}), Sequential Neural Ratio Estimation-C (SNRE-C, \citet{miller2022contrastive}), Sequential Neural Approximate Sufficient Statistics (SNASS, \citet{chen2021neural} and Sequential Neural Approximate Slice Sufficient Statistics (SNASS, \citet{chen2023learning}) on seven synthetic experiments, a solar dynamo model from the astrophysics literature and a neural mass model from neuroscience to highlight the advantages and disadvantages of the method. We chose SNPE-C as neural posterior method since we found it is still state-of-the-art or at least highly competitive on a large number of experimental benchmarks (see, e.g., \citet{deistler2022truncated, wildberger2023flow}). Similarly, SNRE-C is to our knowledge state-of-the-art among methods for neural ratio estimation. Conceptually related to our method, SNASS and SNASSS first compute a set of near-sufficient summary statistics using embedding networks and then use SNL to fit a posterior approximation.

We followed the experimental details of \citet{papamakarios2019sequential}, \citet{greenberg2019automatic} and \citet{miller2022contrastive}. In short, for SSNL, SNL and SNPE-C we use masked autoregressive flows (MAFs) with five flow layers where each layer uses a neural network with two hidden layers and $50$ nodes per layer. Since SNASS and SNASSS have to fit additional summary and critic networks, these use MAFs with three and two layers, respectively, to have roughly the same number of parameters as the previous methods. SSNL uses a dimensionality-reducing surjection in the middle layer for which the conditional density $p\left(z_{k}^- | f^{-1}_k(z_{k}^+; z_{k}^-, \theta), \theta \right)$ is parameterized using an MLP with two layers of $50$ nodes each. We selected the third layer as surjection such that the entire data set is "processed" once in each direction before reducing the dimensionality of the data. We evaluated surjection layers that reduce the dimensionality by $25\%$, $50\%$ or $75\%$, respectively (see Appendix~\ref{appendix:experiment-details} for all experimental details). 

For each experimental model, we sample a vector of true parameters $\theta_\text{obs} \sim p(\theta)$ and then simulate an observation $y_\text{obs} \leftarrow sim(\theta_\text{obs})$ which is then used to approximate $p(\theta | y_\text{obs})$. We repeat this data generating process for $10$ different seeds. We evaluate each method sequentially in $R=15$ rounds using a total simulation budget of $N = 15\ 000$: in each round $r$ we draw a sample $\theta^r_n$ of size $N_R=1\ 000$ from the trained surrogate posterior (or prior if in the first round, respectively), simulate observations $y^r_n \leftarrow {sim}(\theta^r_n)$, and train the density estimator/classifier on all available data (i.e., including the data from all previous rounds, yielding a simulation budget of $1\ 000$ for the first round, $2\ 000$ for the second round, etc.). After training, we compare samples from the posterior approximation of a method of each round to samples obtained from MCMC and compute divergence measures between the two samples. For the solar dynamo and neural mass models, we compare the surrogate posterior samples to the true parameter values $\theta_\text{obs}$ as in prior work (e.g., \citet{rodrigues2021hnpe} or \citet{buckwar2020spectral}).

\subsection{Comparing posterior distributions}
Previous work has evaluated the accuracy of the approximated posteriors to the true posterior (or rather the posterior obtained via Monte Carlo samples), mainly using maximum mean discrepancy (MMD; \citet{gretton2012kernel,sutherland2017generative}) and classifier two-sample tests (C2ST; \citet{lopezpaz2017revisiting}). Recently, \citet{zhao2022comparing} introduced a general H-divergence to assess the similarity of two (empirical) distributions, $p$ and $q$, and demonstrated that their method has higher power than members of the MMD and C2ST families in several experimental evaluations while having a low number of hyperparameters to optimize. Specifically, \citet{zhao2022comparing} propose to use the divergence
\begin{equation*}
    D_\ell^\phi(p || q) = \phi\left(  H_\ell \left(\tfrac{p + q}{2}  \right) - H_\ell(p),
    H_\ell\left(\tfrac{p + q}{2}  \right) - H_\ell(q) \right)
\end{equation*}
where the H-min divergence $D^{\text{Min}}_\ell = H_\ell \left (\tfrac{p + q}{2}  \right) - \text{min}(H_\ell(p), H_\ell(q))$ and H-Jensen Shannon divergence $D^{\text{JS}}_\ell= H_\ell \left (\tfrac{p + q}{2}  \right) - \frac{1}{2}(H_\ell(p), H_\ell(q))$ are special cases. $H_\ell(p) = \text{inf}_{a \in \mathcal{A}} \mathbb{E}_p[ \ell(X, a)]$ is the Bayes optimal loss of some decision function over an action set $\mathcal{A}$ and the loss $\ell$ can in practice be implemented, e.g., using the negative log-likelihood of a density estimator such as kernel density estimator or Gaussian mixture model (see Appendix~\ref{appendix:experiment-details} for details on H-divergences).

We evaluated H-Min and H-Jensen Shannon divergences on the notorious simple-likelihood-complex-posterior model (SLCP; see Appendix~\ref{appendix:model-details} for a description) following the experimental details in \citet{zhao2022comparing} and compared them to MMD and C2ST distances (Figure~\ref{fig:slcp_profiles}). We found that the profiles of H-Min or H-Jensen Shannon have similar trends as C2ST and MMD, respectively (the H-Jensen Shannon divergence is in fact strictly larger than the family of MMD distances \citep{zhao2022comparing}). 

Hence, we propose to use both the H-Min and H-Jensen Shannon divergences as model evaluation metrics for SBI benchmarks due to their power, implementational simplicity and low number of tunable hyperparameters, and will report them for the experimental evaluations. Note that in the SLCP example, which we used to assess the different divergences, SSNL outperforms the three baselines consistently with a sufficient simulation budget.

\subsection{SBI model benchmarks}
We first evaluate SSNL on multiple benchmark models from the SBI literature (i.e., Ornstein-Uhlenbeck, Lotka-Volterra, SIR, and generalized linear model (GLM)) and discuss when it should have performance benefits over alternative methods. We then demonstrate using two negative examples where SSNL breaks and where it should fail to outperform the baselines (Gaussian mixture model and hyperboloid model). For a detailed description of the six experimental models which we omit here, we refer to  Appendix~\ref{appendix:model-details}.

\begin{figure}
\begin{center}
\subfloat[Posterior divergences.]{
    \label{fig:synthetic_model_benchmarks-a}
    \includegraphics[width=\columnwidth]{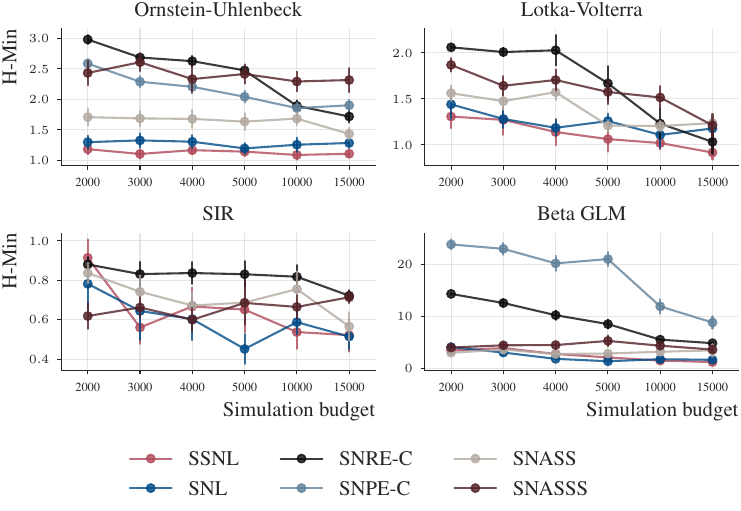}
}

\subfloat[Likelihood profiles.]{    
    \label{fig:synthetic_model_benchmarks-b}
    \includegraphics[width=\columnwidth]{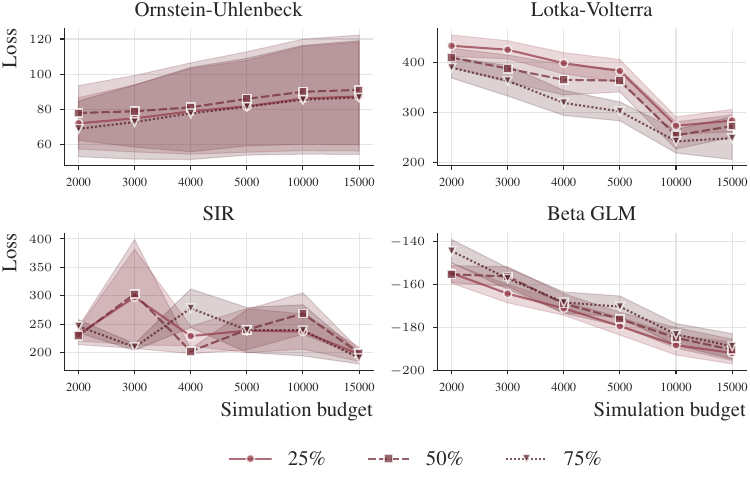}
}

\subfloat[Autocorrelations and intrinsic dimensions.]{
    \label{fig:synthetic_model_benchmarks-c}
    \includegraphics[width=\columnwidth]{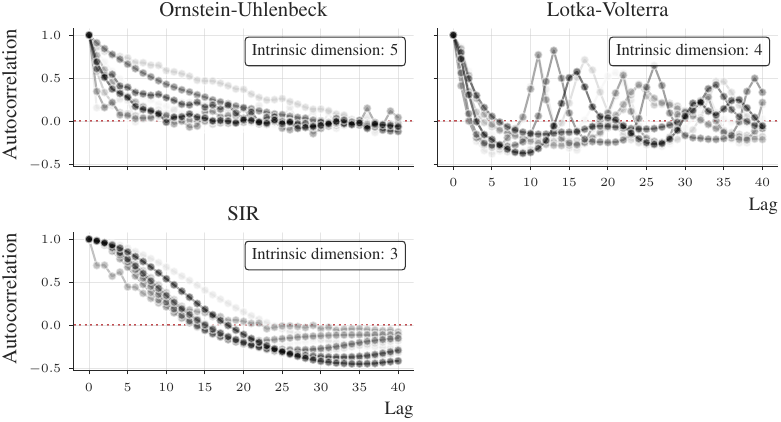}
}
\caption{Experimental results of OU, LV, SIR and GLM models. (a) H-Min divergences of all models plotted against the size of simulated data (lower is better). (b) Validation likelihood profiles of SSNL models when the middle layer reduces the dimensionality by $25\%$, $50\%$ or $75\%$, respectively. The performances are similar for all models which is why we used the most conservative reduction, i.e. $75\%$ for all models. (c) Autocorrelation (AR) plots for the three time series models up to a lag of $40$ (black shades correspond to realizations of a time series with different parameter values). The AR for the OU model converges to zero while the AR for the LV has a self-repeating structure. The SIR model has a single saddle-point and does not converge.}
\label{fig:synthetic_model_benchmarks}
\end{center}
\vskip -0.2in
\end{figure}

\paragraph{Results} For SSNL, we first determined the optimal embedding dimensionality in the following way: we extract the validation loss, i.e., the negative log-likelihood on the validation set, after training and use the embedding dimensionality corresponding to the network that achieved the lowest validation loss (Figure~\ref{fig:synthetic_model_benchmarks-b})\footnote{This could be done more rigorously, e.g., by splitting the data into an additional test set and evaluating its loss or by simply reducing the dimensionality sequentially such that the embedding has as many dimensions as required summary statistics, but we found this simple heuristic to be sufficient and it does not require additional computation. Furthermore, since the data is simulated with iid noise, the likelihoods on validation and test sets should be almost equal. Alternatively, information-theoretical approaches could also be applied.}. Since the loss profiles on all four experiments are roughly the same for each parameterization, we, for simplicity, chose to use the networks that reduce the dimensionality to $75\%$ for each experimental model.

For the two time series models Ornstein-Uhlenbeck (OU) and Lotka-Volterra (LV), SSNL consistently outperforms all baselines. SSNL is on par with SNL on the SIR and Beta GLM models (see Figure~\ref{fig:synthetic_model_benchmarks-a}). The SIR model is the only case with mixed, inconsistent results where for different simulation budgets SNL gets outperformed by SSNL or vice versa (the figures do not show SNPE-C for LV and SIR due to its bad performance, see Figure~\ref{app:fig-all-four-benchmarks-results} in the Appendix for complete results). 

\paragraph{Autocorrelation and intrinsic dimensionality}
We assessed in which case and why SSNL has a performance advantage over SNL and argue that a combination of autocorrelation and intrinsic dimensionality (ID) of a data set might be indicative of it (see Figure~\ref{fig:synthetic_model_benchmarks-c} where realizations of a time series model with different parameter values are shown). Notably, for the Ornstein-Uhlenbeck and Lotka-Volterra models the autocorrelation pattern seems to benefit dimensionality-reducing methods. For the Ornstein-Uhlenbeck process, the autocorrelation converges to zero when considering longer lags meaning that information beyond a certain point is not informative of the parameters any more. Similarly, for the Lotka-Volterra process the autocorrelation patterns are repetitive after a certain lag meaning that the data at larger time points is basically a copy of previous time points (compare Figure~\ref{fig:sbi_benchmark_data_visualisation} in the Appendix). In the case of the SIR model, the autocorrelation has first an negative slope and then changes the sign of its gradient function after reaching a saddle-point. Consequently, the entire time-series is informative of the parameters and dimensionality reduction has supposedly only little advantage over dimension-preservation (more experimental results can be found in Appendix~\ref{appendix:additional-results}). 
\begin{figure}
\begin{center}
\includegraphics[width=\columnwidth]{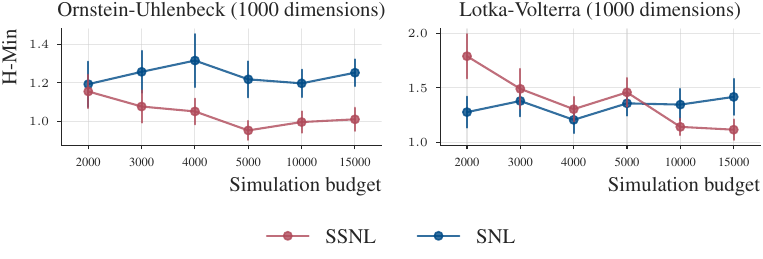}
\caption{Posterior divergences on $1000$-dimensional Ornstein-Uhlenbeck and Lotka-Volterra models.}
\label{fig:synthetic_model_benchmarks_long_timeseries}
\end{center}
\end{figure}

\paragraph{Increasing the dimensionality} The benefit of dimension reduction depends arguably on the length of the time series and its signal-to-noise ratio. To validate this hypothesis, we replicated the OU and LV experiments but increased the number of time points from $100$ to $1000$ and observed that the performance difference between SSNL and SNL in fact increases. For Ornstein-Uhlenbeck, SSNL has a significant performance advantage over SNL, while for Lotka-Volterra the same can be observed with sufficient simulation budget (Figure~\ref{fig:synthetic_model_benchmarks_long_timeseries}). We hypothesize that this is due to the fact that in some cases learning the high-dimensional conditional density $p\left(y_-| f^{-1}(y_+; y_-, \theta)\right)$ requires an increased sample size.

\paragraph{Negative examples and limitations} The performance of SSNL depends on whether the parameter-related information in the data can be represented in a lower-dimensional space. In scenarios where this is not the case, e.g., on Gaussian mixture or hyperboloid models, SNPE-C or SNL expectedly outperform SSNL (see Figure~\ref{fig:negative-examples} in the Appendix).

\subsection{Solar dynamo}
We applied SSNL to a real-world solar dynamo model from the solar physics literature that models the magnetic field strength of the sun (see \citet{charbonneau2005fluctuations} and references therein). The model is a non-linear time series model with both additive and multiplicative noise terms
\begin{align*}
g(y) &= \tfrac{1}{2} [1 + \text{erf}( \tfrac{y  - b_1}{w_1})] [1 - \text{erf} (\tfrac{y  - b_2}{w_2} ) ] \\ 
 \alpha_t &\sim \mathcal{U}(\theta_1, \theta_1 + \theta_2), \quad \epsilon_t \sim \mathcal{U}(0, \theta_3), \\
y_{t + 1} &\leftarrow \alpha_t g(y_t) y_t  + \epsilon_t
\end{align*}
The model is interesting, because it has more noise components than observed outcomes and integrating out the noise components yields a marginal likelihood that is outside the exponential family. Consequently, the number of sufficient statistics for such a model is unbounded (with the length of the time series $T$) according to Pitman-Koopman-Darmois theorem. We choose the prior $p(\theta)$ and hyperparameters $b$ and $w$ as in \citet{albert2022learning} and simulate a single time series of length $T=100$ (see Appendix~\ref{appendix:experiment-details-solardynamo} for details).

\begin{figure}
\begin{center}
\subfloat[Solar dynamo]{
\includegraphics[width=0.95\columnwidth]{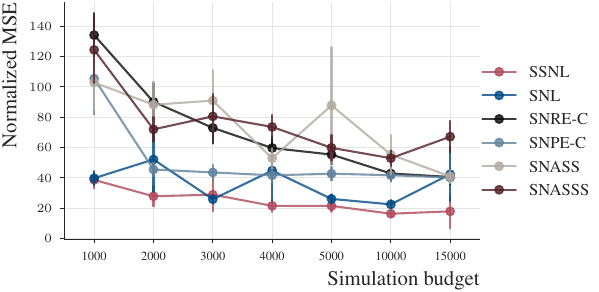}
\label{fig:exp-models-validation-sd}%
}
\newline
\subfloat[Neural mass model]{%
\includegraphics[width=0.95\columnwidth]{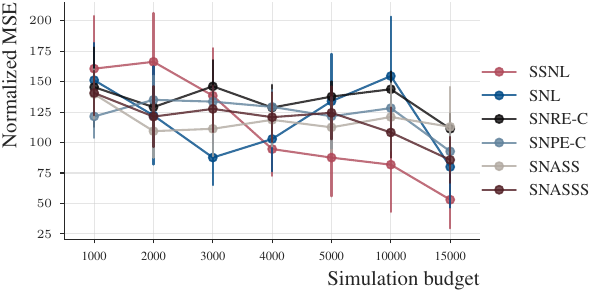}
\label{fig:exp-models-validation-jr}%
}
\caption{Solar dynamo and neural mass model evaluation (we show the MSE w.r.t. the prior sample $\theta_\text{obs}$ that was used to simulate the observation $y_\text{obs}$. Values are normalized by the minimum MSE in the entire data set).}
\label{fig:exp-models-validation}
\end{center}
\end{figure}

\begin{figure*}
\begin{center}
\includegraphics[width=\textwidth]{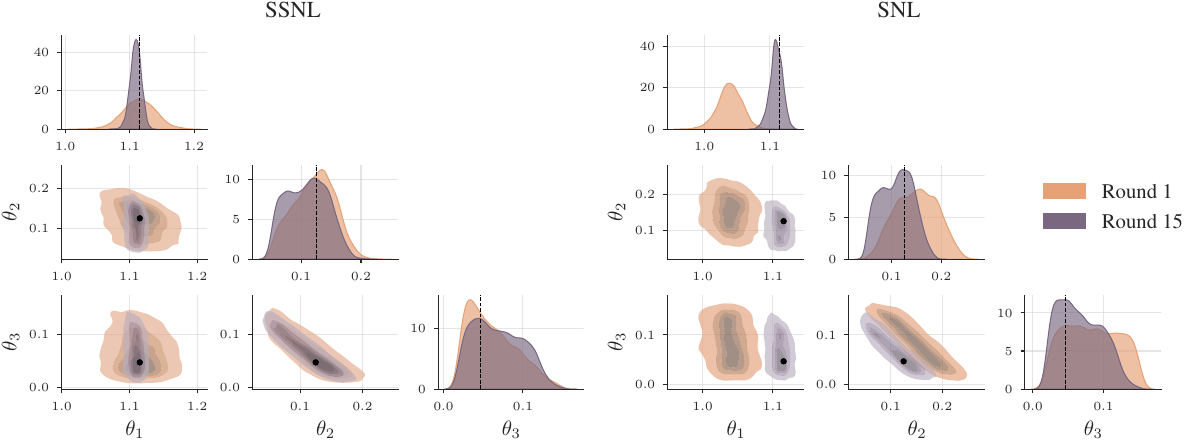}
\caption{Solar dynamo posterior distributions of SSNL and SNL after the 1st and 15th round (shown as kernel density estimates. Black dots and lines represent true parameter values).}
\label{fig:solardynamo-posteriors}
\end{center}
\end{figure*}

SSNL consistently outperforms the five baselines for this experiment (Figure~\ref{fig:exp-models-validation-sd} left column). Having a closer look at the posterior distributions of one experimental run, one can observe that SSNL already after the first round recovers the true parameters reliably while the posterior mean of SNL is heavily biased (Figure~\ref{fig:solardynamo-posteriors}). After the final round, both methods converge to the true parameter values.

\subsection{Neural mass model}
We also evaluate SSNL on the stochastic version of the Jansen-Rit neural mass model \citep{ableidinger2017stochastic} which describes the collective electrical activity of neurons by modelling interactions of cells (see \citet{ableidinger2017stochastic,buckwar2020spectral,rodrigues2021hnpe} for details). The model is a $6$-dimensional SDE of the form
\begin{equation*}
\begin{split}
\mathrm{d} \begin{pmatrix}
R_t\\
S_t
\end{pmatrix} =
\begin{pmatrix}
S_t\\
-\Gamma^2R_t - 2\Gamma S_t + G_\theta(R_t)
\end{pmatrix}
\mathrm{d}t
+ \begin{pmatrix}
0\\
\Sigma_\theta
\end{pmatrix}
\mathrm{d}W_t
\end{split}
\end{equation*}
where $R = [Y_1, Y_2, Y_3]^T$, $S = [Y_4, Y_5, Y_6]^T$, $W_t$ is a Wiener process, $\Sigma_\theta$ is a diagonal covariance matrix, $G_\theta$ is a displacement vector, $\Gamma$ is a matrix, and $\theta$ is a four-dimensional random vector with uniform prior (see Appendix~\ref{appendix:experiment-details-jansenrit} for details). 

With a sufficient simulation budget, in this case $4000$ simulations, SSNL convincingly outperforms the baselines having the lowest MSE. As before (see, e.g., Figure~\ref{fig:synthetic_model_benchmarks_long_timeseries}), we hypothesize that the performance of SSNL is worse for lower simulation budgets, since an additional conditional density has to be learned. Intriguingly, the inferences of SNL, which like SSNL approximates the likelihood function, are significantly worse than for SSNL and overall inconsistent, indicating that SSNL is in general an excellent off-the-shelf estimator for high-dimensional data sets (more experimental results can be found in Appendix~\ref{appendix:additional-results}).

\section{{Related work}}
\label{sec:related}
Neural simulation-based inference is an emergent field with significant advances in the recent past. In addition to (sequential) neural approximations to the posterior \citep{papamakarios2016fast,lueckmann2017flexible,greenberg2019automatic,deistler2022truncated,wildberger2023flow}, likelihood \citep{papamakarios2019sequential,glockler2022variational} or likelihood-ratio \citep{cranmer2015approximating,durkan2020contrastive,hermans2020likelihood,thomas2022likelihood,delaunoy2022towards,miller2022contrastive}, recent approaches have utilized flow matching and score-based models for posterior inference \citep{schmitt2023consistency,geffner2023compositional,wildberger2023flow}, albeit mainly in a non-sequential manner. Score-based NPE methods are a fruitful research direction which do not restrict the architecture of the neural network architecture. More recently, \citet{gruner2023pseudolikelihood} proposed a new method that is targeted at models in which the posterior is conditioned on multiple observations simultaneously. \citet{jia2024simulation} introduce a new family of methods, called \textit{neural quantile estimation}, which uses quantile regression to either approximate the intractable posterior or likelihood functions of a model. \citet{glaser2022maximum,pacchiardi2022score} propose approaches to SBI using energy-based models as density estimators. Finally, \citet{yao2023simulation} propose a method based on stacking to combine the results of multiple posterior inferences, i.e., when multiple posterior approximations from different methods are available. 

Related to our work, \citet{alsing2019nuisance,chen2021neural,chen2023learning} have developed methods for SBI to compute summary statistics which, however, requires learning an additional embedding network while our approach learns an embedding and computes likelihood approximations in a single step and without computational overhead. \citet{radev2023jana} discuss an approach that learns three networks for posterior approximations, likelihood approximations and summary statistic computation. \citet{beck2022efficient} discuss marginalization of data dimensions to see the influence of different covariates on the posterior distribution and propose an approach to find informative data dimensions.

In order to evaluate inferences of non-sequential procedures of models for which samples from the true posterior are not available, \citet{linhart2023lcst} have proposed a local procedure based on classifier tests. Similarly, \citet{yao2023discriminative} proposed a diagnostic related to, and with higher power than, simulation-based calibration (\citet{talts2018validating}; unfortunately, for neural likelihood methods we found them computationally infeasible to use, since they require either computing a vast number of permutation tests of which each requires learning a classifier or repeatedly sampling from the surrogate posterior for every model).

\section{{Conclusion}}
\label{sec:conclusion}
We introduced \textit{Surjective Sequential Neural Likelihood} estimation, a new neural likelihood method for simulation-based inference for high-dimensional data. SSNL uses dimensionality-reducing surjections to embed the data in a lower-dimensional space while simultaneously learning the likelihood function.

SSNL performs particularly well when applied to high-dimensional time series data outperforming state-of-the-art methods and is on par with them in other experiments, making it an excellent off-the-shelf estimator for high-dimensional data sets. As a limitation, we identified that due to introducing an additional conditional density, SSNL in some cases requires a higher sample size than other flow-based models like SNL or SNPE-C.

We anticipate that our method will complement other approaches in SBI well, but do not argue that it should be generally preferred over others. For instance, when posteriors have a simple geometry and the likelihood function is complicated to approximate, posterior estimators like SNPE-C often provide superior performance. Similarly, neural ratio methods, such as SNRE-C, are typically easier to fit as flow-based methods, since they only require training a classifier.

Future research could elucidate how deeper (stacked) surjective architectures impact posterior inference and if more suitable surjective architectures can be developed for SBI. The application of information-theoretic criteria to more rigorously determine the dimensionality of the latent space, for instance, as done in \citet{chen2023learning}, is another possible research direction.

\begin{acknowledgements}
This research was supported by the Swiss National Science Foundation (Grant No. $200021\_208249$). We thank all anonymous reviewers for their feedback which helped to improve the manuscript.
\end{acknowledgements}


\bibliography{references}

\newpage
\onecolumn
\title{Simulation-based Inference for High-dimensional Data using Surjective Sequential Neural Likelihood Estimation\\(Supplementary Material)}
\maketitle

\appendix
\section{Background}
\label{appendix:more-background}

\subsection{Neural posterior estimation}

Neural posterior estimation (SNPE) methods \citep{papamakarios2016fast,lueckmann2017flexible,greenberg2019automatic,deistler2022truncated,wildberger2023flow} use a normalizing flow to directly target the posterior distribution thereby approximating $q_{f}( \theta | {y}) \approx p( \theta | {y})$. SNPE-C \citep{greenberg2019automatic}) uses the same sequential training procedure as SNL. In the first round, however, it optimizes the maximum likelihood objective $\mathbb{E}_{\mathcal{D} } \left[ q_{f}( \theta | {y}) \right]$ in each round. Subsequent rounds proceed by first composing a proposal prior as $\hat{p}^r( \theta) = q_{f}( \theta | {y}_0)$, simulating new pairs $\{ {y}_n, \theta_n \}^r_{1 \dots N}$ where $ \theta_n \sim \hat{p}^r( \theta)$ and then re-training the NF. Since the parameters are sampled from the proposal prior $\hat{p}^r( \theta)$, the surrogate posterior would no longer target the true posterior $p( \theta | {y})$ but rather 
\begin{equation*}
 q_{f}( \theta | {y}) \propto p( \theta | {y}) \frac{\hat{p}^r(\theta)}{p(\theta)}
\end{equation*}
\citet{greenberg2019automatic} overcome this by deriving the new objective $\mathbb{E}_{\mathcal{D} } \left[   \frac{1}{Z} q_{f}( \theta | {y})  \frac{\hat{p}^r(\theta)}{p(\theta)} \right]$ which however requires the computation of a normalization constant $Z$. 

SNPE-C can simulate posterior realizations by sampling from the normalizing flow base distribution first,
and then propagating the samples through the flow layers. This can lead to posteriors that are outside
of the prior bounds which need to be rejected and the procedure repeated until a sample of desired
size has been taken. Specifically, if the prior distributions are constrained, e.g., containing scale
parameters, or are very narrow, APT is known to exert ’leakage’, i.e., the posterior approximation
might produce samples that are not within the prior bounds. In this case, the rejection rate of posterior
samples is elevated, for instance, as reported in \citet{durkan2020contrastive} or \citet{glockler2022variational}, the
latter of which having observed rejection rates of up to $99\%$, which necessitates the use of MCMC
methods instead. Leakage significantly reduces the usefulness of SNPE methods in comparison to
SNL where draws are generated using MCMC in the first place. Furthermore, for structured data sets,
e.g., time series data, SNPE-C requires facilitating a second neural network to embed the data before
conditioning which increases the number of effective parameters.

\subsection{Likelihood ratio estimation}

Neural likelihood ratio estimation (NRE) methods \citep{hermans2020likelihood,durkan2020contrastive,miller2022contrastive,delaunoy2022towards}) learn the likelihood-to-evidence ratio ${r}({y}, \theta) = \frac{p({y} | \theta)}{p({y})} = \frac{p( \theta | y)}{p(\theta)}$ and then build a surrogate posterior $\hat{p}(\theta | {y}) = \hat{r}({y}, \theta) p( \theta)$. A major advantage of NRE methods is, that they do not need to train a model that estimates a density using normalizing flow which often brings significant computational and numerical advantages. While NRE-C \citep{miller2022contrastive} has been proposed in a non-sequential scenario, it is also possible to derive posterior distributions sequentially (SNRE-C; see, e.g., \citet{tejero2020sbi}). In this case, since a proposal posterior $p^r(\theta|y_0)$ is derived after round $r$, which changes the joint distribution to $p(y|\theta)p^r(\theta|y_0)$, the estimated ratio becomes
\begin{equation*}
 {r}({y}, \theta) = \frac{p({y} , \theta)}{p^r(\theta|y_0)} = \frac{p( \theta | y)}{p(\theta)}   
\end{equation*}
Consequently, the true posterior can only be estimated up to a constant:
\begin{equation*}
p(\theta|y) \propto {r}({y}, \theta) p(\theta)
\end{equation*}

\subsection{Notes}

The idea of using non-trivial embedding networks, such as CNNs or LSTMs for NPE and NRE methods is not new (see e.g., \citet{greenberg2019automatic} or the notebooks of the SBI Python package\footnote{\url{https://sbi-dev.github.io/sbi/tutorial/05_embedding_net/}}). This requires an additional neural network and consequently increases the number of total parameters. We, on the other hand, do dimensionality reduction and likelihood estimation in one step and with one network.

\section{Mathematical derivations}
\label{appendix:surjection-layer}

The derivation of the surjection layer used in SSNL largely follows the SurVAE framework of \citet{nielsen2020survae} and \citet{klein2021funnels}. The SurVAE framework models the log-probability $\log p({y})$ of a $P$-dimensional data point ${y} \in \mathcal{Y}$ as

\begin{align}
\log p({y}) = \log p \left( {z} \right) + V({y}, {z}) + E({y}, {z}) , \qquad {z} \sim q({z} | {y})
\label{eqn:survae}
\end{align}

where $q({z} | {y})$ is some amortized (variational) distribution, ${z} \in {Z}$ is a latent variable with distribution $p({z})$, $V({y}, {z})$ is denoted likelihood contribution term and $E({y}, {z})$ is a bound looseness term. 

\citet{nielsen2020survae} define the likelihood contribution for inference surjections as
\begin{equation*}
V({y}, {z}) = \lim\limits_{q({z} | {y}) \rightarrow \delta \left( {z}  - h^{-1}({y}) \right)} \mathbb{E}_{q({z} | {y})}\left[ \log   \frac{p({y} | {z})}{q({z} | {y})} \right]
\end{equation*}
where $p({y} | {z})$ is some generative stochastic transformation, $h^{-1}: \mathcal{Y} \rightarrow \mathcal{Z}$ is an inference surjection and where we for convenience of notation denote with $h: \mathcal{Z} \rightarrow \mathcal{Y}$
a right inverse function to $h^{-1}$. For bijective normalizing flows the bound looseness term equals $E({y}, {z}) = 0$. For surjective normalizing flows, the same is true if a right inverse $h$ exists (i.e., when the stochastic right inverse condition is satisfied).

By observing (see also Appendix~A of \citet{nielsen2020survae} and main manuscript \citet{klein2021funnels}) that the composition of a differentiable function $g$ with a Dirac $\delta$ function and a bijection $f$ is
\begin{equation*}
    \int \delta \left( g \left( {y} \right)  \right) f\left( g\left( {y} \right) \right) \bigg| \det \frac{\partial g \left( {y} \right) }{\partial {y} }     \bigg| \mathrm{d}{y} = \int\delta \left( {u} \right) f \left( {u} \right) \mathrm{d}  {u} 
\end{equation*}
we can conclude that
\begin{equation*}
    \delta \left( g \left( {y} \right)  \right)  = \delta\left({y} - {y}_0 \right) \bigg| \det \frac{\partial g \left( {y} \right) }{\partial {y} }     \bigg|^{-1}_{{y}={y}_0} 
\end{equation*}
where ${y}_0$ is the root of $g$ (which assumes that $f$ has compact support, the root is unique and that the Jacobian is not singular). 

We now define a conditional bijection $f(z; y_-, \theta)$ and its inverse $f^{-1}(y_+; y_-, \theta)$ for any $Q < P$, set $g({y}) = {z} - f^{-1}(y_+; y_-, \theta)$ (which has its root at ${y}_0 = f(z; y_-, \theta)$) and define
\begin{align*}
    q({z} | {y}) &= \delta \left({z} - f^{-1}(y_+; y_-, \theta) \right) \\
              &= \delta \left({y}_+ - f ( z; y_-, \theta) \right) \big|  \det J^{-1}  \big|^{-1}  \\
\end{align*}
where 
\begin{align*}
    J^{-1} = \frac{\partial f^{-1}(y_+; y_-, \theta) }{\partial {y}_+}    \bigg|_{{y}_+=f({z}; y_-, \theta )}
\end{align*}
Using this result and the conditional distribution $p({y} | {z}) =p({y}_- | {z}, \theta)$ the likelihood contribution for a surjection layer becomes 
\begin{align*}
V({y}, {z}) &= \lim \limits_{q({z} | {y}) \rightarrow \delta \left( {z}  - h^{-1}({y}) \right)} \mathbb{E}_{ q({z} | {y})}\left[ \log \frac{p({y} | {z})}{q({z} | {y})} \right] \\
&= \int \delta \left( {z} - f^{-1}({y}_+ ; y_-, \theta) \right) \log \frac{p({y}_- | {z}, \theta)}{\delta\left( {z} - f^{-1}({y}_+; y_-, \theta) \right)} \mathrm{d}{z}  \\
&= \int \delta \left(y^+ - f(z; y_-, \theta)\right) |\det J^{-1}|^{-1} \log \frac{p(y^- | z, \theta)}{ \delta(y^+ - f(z; y_-, \theta)) |\det J^{-1})|^{-1}  }   \mathrm{d}z \\
&= \int \delta \left(y^+ - \tilde{y}^+\right) \log \frac{p(y^- | z, \theta)}{ \delta(y^T - \tilde{y}^+) |\det J^{-1}|^{-1}  }  \mathrm{d} \tilde{y}^+ \\
&= \log p \left({y}_- | f^{-1}({y}_{+}; y_-, \theta)\right) - \log \big|  \det J^{-1}  \big|^{-1}
\end{align*}

where we used the change of variables $\tilde{y}^+ = f(z; y_-, \theta)$ yielding $\mathrm{d}\tilde{y}^+ = \mathrm{d}z |\det J^{-1}|^{-1}$. 

\section{Implementation details}
\label{appendix:implemention-details}
Surjection layers can be implemented in a straight-forward manner by extending the bijection layers of conventional machine libraries. Below, we demonstrate the implementation of a conditional affine masked autoregressive surjective flow that uses an affine MAF \citep{papamakarios2017masked}, called \texttt{AffineMaskedAutoregressive} as a super class.

\vskip 1em
\begin{lstlisting}[emph={__init__,AffineMaskedAutoregressiveSurjection,AffineMaskedAutoregressive,_inner_bijector,evidence,_inverse_and_likelihood_contribution}]

@dataclass
class AffineMaskedAutoregressiveSurjection(AffineMaskedAutoregressive):
    n_keep: int
    decoder: Callable
    conditioner: MADE
    
    def _inner_bijector(self):
        # define the bijector 'f'
        return AffineMaskedAutoregressive(self.conditioner)

    def _inverse_and_likelihood_contribution(self, y, x=None, **kwargs):
        # here, we define the subsets by just splitting y after some index
        # in general, we do it as describe it as in the main manuscript
        y_plus, y_minus = y[..., :self.n_keep], y[..., self.n_keep:]
        y_cond = y_minus
        
        if x is not None:
            y_cond = jnp.concatenate([y_cond, x], axis=-1)
        # compute lower-dimensional representation
        z, jac_det = self._inner_bijector().inverse_and_log_det(y_plus, y_cond)

        z_condition = z
        if x is not None:
            z_condition = jnp.concatenate([z, x], axis=-1)
        # compute conditional probability
        lc = self.decoder(z_condition).log_prob(y_minus)

        return z, lc + jac_det
        
\end{lstlisting}

where \texttt{MADE} is a masked autoencoder for density estimation \citep{germain2015made}, \texttt{decoder} corresponds to the conditional density $p(y_-|z, \theta)$.

\section{Experimental details}
\label{appendix:experiment-details}

\subsection{Implementation details}
All models are implemented using the Python packages \texttt{sbijax}, \citep{dirmeier2024simulation}, \texttt{surjectors} \citep{dirmeier2024surjectors}, the SBI toolbox \citep{tejero2020sbi}, and the Deepmind JAX ecosystem \citep{jax2018github,deepmind2020jax}. We simulate data from stochastic differential equations using the package Diffrax \citep{kidger2021on}.

\subsection{Training and sampling}

We trained each model using an Adam optimizer with fixed learning rate of $r=0.0001$ and momentums $b_1=0.9$ and $b_2=0.999$. Each experiment uses a mini-batch size of $100$. The optimizer is run until a maximum of $2000$ epochs is reached or no improvement on a validation set can be observed for $10$ consecutive iterations. The validation set consists of $10\%$ of the entire data set, while the other $90\%$ are used for training. For each round, we start training the neural network from scratch and do not continue from the previously learned state. Each model was trained on a HPC computing cluster using a single node consisting of two 18 core Broadwell CPUs (Intel Xeon E5-2695 v4).

We train each method in $R=15$ rounds. Each round a new set of pairs $\{ (y_n, \theta_n) \}^N_{n-1}$ of size $N$ is generated using draws from the prior $\theta_n \sim p(\theta)$ and simulator $y_n \leftarrow sim(\theta_n)$, and then used for training the density estimators or classifier, respectively.

For SSNL and SNL, we used the No-U-turn sampler \citep{hoffman2014no} from the sampling library BlackJAX to sample from the intermediate and final posterior distributions $p^r( \theta | {y}_0)$ using $4$ chains of a fixed length of $\num{10000}$ each of which the first $\num{5000}$ iterations are discarded as burn-in per chain. For SNPE-C and SNRE-C experiments, we use the slice sampler of the SBI toolbox for sampling (which we do in lieu of rejection sampling to avoid leakage; see the Appendix~\ref{appendix:more-background}).Samples from the "true" posterior distribution have been drawn using TensorFlow Probability's slice sampler where we used $10$ chains of length $\num{20000}$ of which we discarded the first $\num{10000}$ as burn-in. Convergence of the true posteriors in this case has been diagnosed using the potential scale reduction factor \citep{gelman1992inference,vehtari2021rhat}, effective sample size calculations, and conventional graphical diagnostics \citep{gabry2019visualization}.

\subsection{Neural network architectures}

For all experiments and evaluated methods, we used the same neural network architectures. We followed the neural network architectures as described in \citet{greenberg2019automatic,papamakarios2019sequential,miller2022contrastive} and tried to keep the number of total parameters of each model as comparable as possible to allow for an unbiased evaluation.

\paragraph{SSNL} The SSNL architectures use a total of $K=5$ layers, the third of which is a surjection layer with reduction factors of $25\%$, $50\%$ or $75\%$ which we chose arbitrarily. For instance, for a reduction factor of $25\%$, we take the initial dimensionality $P$ and reduce it to $Q = \lfloor 0.25 * P \rfloor$. Each of the layers is parameterized by a MAF which uses a MADE network with two layers with $50$ neurons each as conditioner \citep{germain2015made,papamakarios2017masked}. The MAFs use tanh activation functions. The conditional densities are parameterized using a two-layer MLP with tanh activation functions. Between each MAF layer, we add a permutation layer that reverses the vector dimensions. In practice, we assume that optimizing the number of surjection layers and their reduction factors is advisable. This can, for instance, be done empirically by examining the likelihood profiles during training or by reducing to the same order of magnitude as required summary statistics.

\paragraph{SNL} SNL uses a total of $K=5$ layers. Each of the layers is parameterized by a MAF which uses a MADE network with two layers with $50$ neurons each as conditioner \citep{germain2015made,papamakarios2017masked} with permutations in between which reverse the vector dimensions. SNL uses tanh activation functions.

\paragraph{SNPE-C} SNPE-C uses the same normalizing flow architecture as SNL. We, otherwise, use the default SNPE-C parameterisation of the SBI toolbox which uses $10$ atoms for classification.

\paragraph{SNRE-C} SNRE-C architectures consist of MLP networks with two layers and $50$ nodes per layers. SNRE-C uses ReLU activation functions. We, otherwise, use the default SNRE-C parameterisation of the SBI toolbox which uses $5$ classes to classify against and $\gamma=1.0$.

\paragraph{SNASS} To keep the number of parameters as equal as possible, SNASS uses a normalizing flow using three flow layers each consisting of a MADE with two layers and 50 nodes each. SNASS uses as summary and critic networks two MLPs with two hidden layers and 50 nodes each. All activation functions are tanhs.

\paragraph{SNASSS} Similarly, SNASS uses a normalizing flow using two flow layers each consisting of a MADE with two layers and 50 nodes each. SNASSS uses as summary and critic networks three MLPs with a single hidden layer and 50 nodes. All activation functions are tanhs.

\subsection{Estimation of divergences}

\citet{zhao2022comparing} propose using the H-divergence

\begin{equation*}
    D_\ell^\phi(p || q) = \phi\left(  H_\ell\left(\frac{p + q}{2}  \right) -H_\ell(p),
    H_\ell \left(\frac{p + q}{2}  \right) -H_\ell(q)\right)
\end{equation*}

to compare two empirical distributions. Here, $H_\ell(p) = \text{inf}_{a \in \mathcal{A}} \mathbb{E}_p[\ell(X, a)]$ is the Bayes optimal loss of some decision function over an action set $\mathcal{A}$.

They further illustrate the H-Min divergence
\begin{equation*}
    D^{\text{Min}}_l = H_\ell \left (\frac{p + q}{2}  \right) - \text{min} \left( H_\ell(p), H_\ell(q) \right)
\end{equation*}
and H-Jensen Shannon divergence
\begin{equation*}
    D^{\text{JS}}_\ell = H_\ell \left (\frac{p + q}{2}  \right) - \frac{1}{2} \left(  H_\ell(p), H_\ell(q)  \right)
\end{equation*}
as two special cases.

We compute $H_\ell$ using the negative log-likelihood of kernel density estimators (KDEs) using $5$-fold cross-validation. Specifically, we first fit KDE with Gaussian kernels on samples of size $N=\num{10000}$ from the true posterior distribution $p$ $\mathcal{P} = \{ \theta^{\text{true}}_n \}^{N}_{n=1}$ and surrogate posterior distribution $q$ $\mathcal{Q} = \{ \theta^{\text{surrogate}}_n \}^{N}_{n=1}$ separately (i.e., one KDE for each set of samples). We find the optimal hyperparameters $\phi \in \{ 0.1, \dots, 5 \}$ for each KDE using a grid search. We then, in $F=5$ different iterations (folds), subsample $\mathcal{S}_f \subset \mathcal{S}$
and $\mathcal{T}_f \subset \mathcal{T}$ randomly (i.e., with equal probability of being in the subsample) and estimate a KDE on $\mathcal{F}_f = \mathcal{S}_f \cup \mathcal{T}_f$. We compute estimates of $H_l$ via

\begin{align*}
    H_\ell(p) &\approx \inf_a \frac{1}{N} \ell(\mathcal{P}, a) \\
    H_\ell(q) &\approx \inf_a \frac{1}{N} \ell(\mathcal{Q}, a) \\
    H_\ell^f \left( \frac{p + q}{2} \right) & \approx \inf_a \frac{1}{N} \ell(\mathcal{S}_f, a) \\
\end{align*}

where $\ell( \mathcal{X}, a )$ is the negative log-likelihood of the data set $\mathcal{X}$ given the optimized hyperparameters (action) $a$.

\subsection{Estimation of intrinsic dimensionality}

We computed the intrinsic dimensionality of a data set using the "tight local intrinsic dimensionality estimator" (TLE) algorithm \citep{amsaleg2019intrinsic,amsaleg2022intrinsic}. The TLE is an estimator of the local intrinsic dimensionality, i.e., the intrinsic dimension of each data point in a data set. Mathematically, the local intrinsic dimensionality for a data point $x$ w.r.t to the distance $r := r(x)$ to its $k$ nearest neighbors is defined as

\begin{equation*}
    \text{ID}(x) = \lim_{r\rightarrow 0} \lim_{\epsilon \rightarrow 0}  \frac{\log( F((1 + \epsilon) \cdot r) / F(r))}{\log(1 + \epsilon)}
\end{equation*}

where we denote with $F(r)$ the cdf of $R$ which can be estimated empirically. The intrinsic dimension of a data point $x$ describes the relative rate at which $F(r)$ increases.

For details and the derivation of the TEL intrinsic dimensionality estimator, we refer the reader to \citet{amsaleg2022intrinsic}.

For Figure~\ref{fig:exp-models-validation}c, we simulated $N=\num{1000}$ observations $\{y_n \}_{n=1}^N$ from the generative models of the Ornstein-Uhlenbeck, Lotka-Volterra and SIR models, respectively, and estimated the local intrinsic dimensions of data set using the Python package \texttt{scikit-dimension} \citep{bac2021scikit}. The package contains several different estimators for local intrinsic dimensionality, and we chose the TLE estimator arbitrarily.

\subsection{Source code}

Source code including detailed instructions to reproduce and replicate all experiments can be found in the supplemental material or on GitHub at \href{https://github.com/dirmeier/ssnl}{\texttt{github.com/dirmeier/ssnl}}.

\section{Additional details on experimental models}
\label{appendix:model-details}

This section describes the nine experimental models in more detail.

\subsection{Simple likelihood complex posterior}

The simple likelihood complex posterior (SLCP, \citet{papamakarios2019sequential}) model with $8$ dimensions uses the following generative process
\begin{equation*}
\begin{split}
\theta_i &\sim \text{Uniform}(-3, 3) \; \text{for} \; i=1, \dots, 5\\
\mu( {\theta}) &= (\theta_1, \theta_2), \phi_1 = \theta_3^2 , \phi_2 = \theta_4^2 \\
\Sigma( {\theta} ) &=
\begin{pmatrix}
\phi_1^2 & \text{tanh}(\theta_5) \phi_1 \phi_2 \\
\text{tanh}(\theta_5) \phi_1 \phi_2 & \phi_2^2
\end{pmatrix}\\
{y}_j | {\theta}  &\sim \mathcal{N}({y}_j; \mu( {\theta}), \Sigma( {\theta})) \; \text{for} \; j=1, \dots, 4\\
{y} &= [{y}_1, \dots, {y}_4]^T
\end{split}
\end{equation*}
The SLCP generally favours neural likelihood methods of neural posterior methods, since modelling a simple likelihood and then sampling from a multi-model posterior is easier in comparison to vice versa. For each round $r$, we generated $1000$ pairs $\{ (y_n, \theta_n)\}$ from the SLCP model.

\subsection{Ornstein-Uhlenbeck}

The Ornstein-Uhlenbeck (OU) process \citep{sarkka2019applied} is a one-dimensional stochastic differential equation that models velocity of a particle suspended in a medium. It has the following form:
\begin{equation*}
dY_t = - \theta_2 (Y_t - \theta_1) \mathrm{d}t + \theta_3 \mathrm{d}W_t
\end{equation*}
where $W_t$ is a Wiener process and $\theta$ are the parameters of interest. The presentation of the OR process above uses an additional drift term $\theta_2$.

Instead of solving this SDE numerically, the OR process admits the analytical forms
\begin{equation*}
Y_t \mid Y_0 = y_0 \sim \mathcal{N}\left( \theta_1 + (y_0 - \theta_1) e^{- \theta_2 t}, \tfrac{\theta_3^2}{2 \theta_2} (1 - e^{-2\theta_2 t})   \right)
\end{equation*}
and
\begin{equation*}
Y_t \mid Y_{s} = y_{s} \sim \mathcal{N}\left( \theta_1 + (y_0 - \theta_1) e^{- \theta_2 (t - s)}, \tfrac{\theta_3^2}{2 \theta_2} (1 - e^{-2\theta_2 (t - s)})   \right)
\end{equation*}
where $s < t$. The conditional density above can be used both for sampling and evaluating the density of an observation $Y_t$. We simulate the OU process for each experiment using the generative model
\begin{align*}
\theta_1 & \sim \mathcal{U}(0, 10) \\ 
\theta_2 &\sim \mathcal{U}(0, 5) \\
\theta_3 & \sim \mathcal{U}(0, 2) \\
Y_t \mid Y_{s} = y_{s} & \sim \mathcal{N}\left( \theta_1 + (y_0 - \theta_1) e^{- \theta_2 (t - s)}, \tfrac{\theta_3^2}{2 \theta_2} (1 - e^{-2\theta_2 (t - s)})   \right)
\end{align*}
and initialize $y_0 = 0$. We sample $100$ equally-spaced observations $Y_t$ where $t \in \{0, \dots, 10 \}$. The conditional density can be used both for sampling and evaluating the density of an observation $Y_t$. 

\subsection{Lotka-Volterra}

The Lotka-Volterra model is a model from ecology that describes the dynamics of a "prey" population and a "predator" population:
\begin{equation*}
\begin{split}
\theta_1 &\sim \text{LogNormal}(-0.125, 0.5) \\
\theta_2 &\sim \text{LogNormal}(-3, 0.5) \\
\theta_3 &\sim \text{LogNormal}(-0.125, 0.5) \\ 
\theta_4 &\sim \text{LogNormal}(-3, 0.5) \\
\tfrac{\mathrm{d}X_1}{\mathrm{d}t} &= \theta_1 X_1 - \theta_2 X_1 X_2 \\
\tfrac{\mathrm{d}X_2}{\mathrm{d}t} &= - \theta_3 X_1 + \theta_4 X_1 X_2 \\
(Y_{t1}, Y_{t2}) &\sim \text{LogNormal}\left(\log \left(X_{t1}, X_{t2}\right), 0.1 \right) \\
\end{split}
\end{equation*}
where $X_1$ are is the density of the prey population and $X_2$ is the density of some predator population. The parameter $\theta = [\theta_1, \dots, \theta_4]^T$ describes growth and death rates, respectively, and effects of presence of predators and prey, respectively.

We follow the parameterization in \citet{lueckmann2021benchmarking}, but sample a longer time series, i.e., $50$ equally-spaced observations $Y_t$ where $t \in [0, 30]$. We then concatenate the two $50$-dimensional vectors $y_t = [y_{t1}^T, y_{t2}^T]^T$ yielding a $100$-dimensional observation.

We solve the ODE using the Python package Diffrax using a Tsit5 solver.

\subsection{SIR model}

The SIR model is a model from epidemiology that describes the dynamics of the number of individuals in three compartmental states (susceptible, infectious, or recovered) which is, for instance, be aplpied to model the spread of diseases. We again adopt the presentation by \citet{lueckmann2021benchmarking} which defines the generative model
\begin{equation*}
\begin{split}
\theta_1 &\sim \text{LogNormal}(\log(0.4), 0.5) \\
\theta_2 &\sim \text{LogNormal}(\log(1/8), 0.2) \\
\frac{\mathrm{d}S}{\mathrm{d}t} & = -\theta_1 \frac{SI}{N}\\
\frac{\mathrm{d}I}{\mathrm{d}t} & = \theta_1 \frac{SI}{N} - \theta_2 I\\
\frac{\mathrm{d}R}{\mathrm{d}t} & = \theta_2 I\\
Y_t & \sim \text{Binomial}\left(1000, \frac{I_t}{N}\right) 
\end{split}
\end{equation*}
where we set $N = \num{1000000}$, and the initial conditions $s_0= N-1$, $i_0=1$ and $r_0=0$. We sample 100 evenly-spaced observations $Y_0$ where $t \in \{0, \dots, 160 \}$. 

Previous work, e.g., \citet{lueckmann2021benchmarking}, used continuous normalizing flows (i.e., pushforwards from a continuous base distribution, not continuous normalizing flows as in \citet{chen2018cnf,grathwohl2019scalable}). Continuous likelihood-based models, such as SNL using MAFs, cannot adequately represent discrete data. As a remedy, we dequantize the counts uniformly after sampling them \citep{Theis2016a}, i.e.,  we add noise $u_t \sim \mathcal{U}(0, 1)$, such that $\tilde{y}_t = y_t + u_t$ and use the noised data for trained. While other approaches to dequantization, such as \citet{ho2019flow}, would possibly be more rigorous, we found that this simple approach suffices.

We solve the ODE using the Python package Diffrax using a Tsit5 solver.

\subsection{Beta generalized linear model}

We evaluated SSNL and the three baselines against a Beta generalized linear regression model (Beta GLM). We use the following generative model
\begin{align*}
\theta & \sim \mathcal{N}(0, B) \\
\eta & = X \theta, \qquad \mu = \text{sigmoid}(\eta)\\ 
Y & \sim \text{Beta}(\mu c, (1 - \mu)c)
\end{align*}
where $c \in \mathbb{R}^+$ is a non-negative concentration parameter, $B$ is computed as in Appendix~T.6 of \citet{lueckmann2021benchmarking}, $\text{Beta}$ describes a Beta distribution with a mean-concentration parameterization \citep{ferrari2004beta} 

In \citet{lueckmann2017flexible}, a Bernoulli GLM is used instead of a Beta GLM(called Bernoulli GLM raw). We changed the Bernoulli likelihood to a Beta likelihood because of the the same reason as for the Lotka-Volterra model (a CNF can generally not model a discrete distribution). We followed the implemtation in SBIBM (\url{https://github.com/sbi-benchmark/sbibm}) as a design matrix $B$.

\subsection{Gaussian mixture model}

The Gaussian mixture (GGM) described in "Negative examples and limitations" in Section~\ref{sec:experiments} uses the following generative process:

\begin{align*}
\theta & \sim \mathcal{U}(-10, 10) \\
Y \mid \theta & \sim 0.5 \mathcal{N}(\theta, I) + 0.5 \mathcal{N}( \theta, \sigma^2 I) 
\end{align*}

where $\sigma^2 = 0.01$ $I$ is a unit matrix, and both $\theta \in \mathbb{R}^2$ and $Y \in \mathbb{R}^2$ are two-dimensional random variables. The GGM again follows the representation in \citet{lueckmann2021benchmarking}.

\subsection{Hyperboloid}

The hyperboloid model \citep{forbes2022summary} described in "Negative examples and limitations" in Section~\ref{sec:experiments} is a 2-component mixture model of $t$-distributions of the form

\begin{align*}
\theta &\sim \mathcal{U}(-2, 2) \\
Y \mid \theta &\sim 
    \frac{1}{2} t(\nu, F(\theta; a) \mathbb{I}, \sigma^2 I) +
    \frac{1}{2} t(\nu, F(\theta; b) \mathbb{I}, \sigma^2 I)
\end{align*}

where $t$ represents a Student's $t$-distribution with $\nu$ degrees of freedom, mean $F(\theta; x) = \left(||\theta - x_1 ||_2 - ||\theta - x_2 ||_2 \right)$ and scale matrix $\sigma^2 I$ and $\mathbb{I}$ is vector of ones. We follow \citet{forbes2022summary}, and in our experiments set $\theta \in \mathbb{R}^2$ to be uniformly distribution, $a_1 = [-0.5, 0.0]^T$, $a_2 = [0.5, 0.0]^T$,
$b_1 = [0.0, -0.5]^T$, $a_2 = [0.0, 0.5]^T$, $\nu = 3$ and $\sigma^2 = 0.01$.

\subsection{Solar dynamo}
\label{appendix:experiment-details-solardynamo}
The solar dynamo model is a non-linear time series model with both additive and multiplicative noise terms
\begin{align*}
\theta_1 &\sim \mathcal{U}(0.9, 1.4) \\
\theta_2 &\sim \mathcal{U}(0.05, 0.25) \\
\theta_3 &\sim \mathcal{U}(0.02, 0.15) \\
g(y) &= \frac{1}{2} [1 + \text{erf}( \tfrac{y  - b_1}{w_1})] [1 - \text{erf} (\tfrac{y  - b_2}{w_2} ) ] \\ 
\alpha_i & \sim \mathcal{U}(\theta_1, \theta_1 + \theta_2) \\
\epsilon_i & \sim \mathcal{U}(0, \theta_3)\\
y_{t + 1} &\leftarrow \alpha_i g(y_t) y_t  + \epsilon_i
\end{align*}

where $\text{erf}(x) = \frac{2}{\sqrt{\pi}} \int_0^x \exp(-t^2) \mathrm{d}y$ is the Gauss error function. We simulate a time series of length $N=100$ recursively starting from $y_0 = [1, 1]^T$. We follow \citet{albert2022learning}, and set $b_1=0.6$, $w_1=0.2$, $b_2=1$ and $w_2=0.8$.

\subsection{Neural mass model}
\label{appendix:experiment-details-jansenrit}

The stochastic version of the Jansen-Rit neural mass model \citep{ableidinger2017stochastic} describes the collective electrical activity of neurons. The model is a $6$-dimensional stochastic differential equation of the form

\begin{align*}
\theta_1 &\sim \mathcal{U}(10, 250)\\
\theta_2 &\sim\mathcal{U}(50, 500) \\
\theta_3 &\sim\mathcal{U}(100, 5000) \\
\theta_4 &\sim \mathcal{U}(-20, 20)\\
\mathrm{d} \begin{pmatrix}
Q_t\\
P_t
\end{pmatrix}  &=
\begin{pmatrix}
P_t\\
-\Gamma^2Q_t - 2\Gamma P_t + G_\theta(Q_t, \theta)
\end{pmatrix}
\mathrm{d}t
+ \begin{pmatrix}
0\\
\Sigma_\theta
\end{pmatrix}
\mathrm{d}W_t
\end{align*}

The actual signal $Y = 10^{g/10} (X_{t1} - X_{t2})$ where $Q = [Y_1, Y_2, Y_3]^T$, $P = [Y_4, Y_5, Y_6]^T$ and $W_t$ is a Wiener process.$\Sigma_\theta = \text{diag}(\sigma_4, \sigma_5,\sigma_6)$ and
$\Gamma = \text{diag}(a, a, b)$ are diagonal $3 \times 3$ matrices with positive $a$ and $b$. The vector

\begin{align*}
G(Q_t; \theta) = \begin{pmatrix}
Aa[\text{sig}(X_2 - X_3)]\\
Aa[\mu + C_2 \text{sig}(C_1 X_1)]\\
Bb[C_4\text{sig}(C_3 X_1)]\\
\end{pmatrix}
\end{align*}

is a $3$-dimensional vector of displacement terms and 

\begin{equation*}
\text{sig}(x) = \frac{v_{\text{max}}}{1 + \exp(r(v_0 - x))}    
\end{equation*}

We are interest in inference of the $4$-dimensional vector $\theta = [\theta_1, \dots, \theta_4]^T = [C, \mu, \sigma, g]^T$. The parameters $C_i$ are related via $C_1 = \theta_1$, $C_2 = 0.8 \theta_1$, $C_3 = 0.25$ and $C_4 = 0.25 \theta_1$. The other parameters are $\mu_4 = \theta_2$, $\sigma_5 = \theta_3$ and $g  = \theta_4$. Following previous work, we initialize $y_0 = [0.08, 18, 15, -0.5, 0, 0]^T$ and simulate a time series $Y_t$ with $t \in [0, 8]$ with sampling frequency $Hz=512$. We then takes $100$ equally-spaced elements from $Y_t$.

We refer the reader to \citet{ableidinger2017stochastic}, \citet{rodrigues2021hnpe} and \citet{buckwar2020spectral} for detailed explanations of all constants and equations from where we also adopted the parameterization: $A=3.25$, $B=22$, $a=100$, $b=50$, $v_{\text{max}} = 5$, $v_0  = 6$, $r = 0.56$, $\sigma_4 = 0.01$ and $\sigma_6 = 1$ (see also \citet{linhart2023lcst}).

We solve the SDE with the Python library \texttt{jrnmm} using the Strang-splitting method as described in \citet{buckwar2020spectral}.

\section{Additional results}
\label{appendix:additional-results}

This section presents additional results for the benchmark models.

\subsection{Four SBI models benchmark}

\begin{figure}[h!]
\begin{center}
\includegraphics[width=0.65\columnwidth]{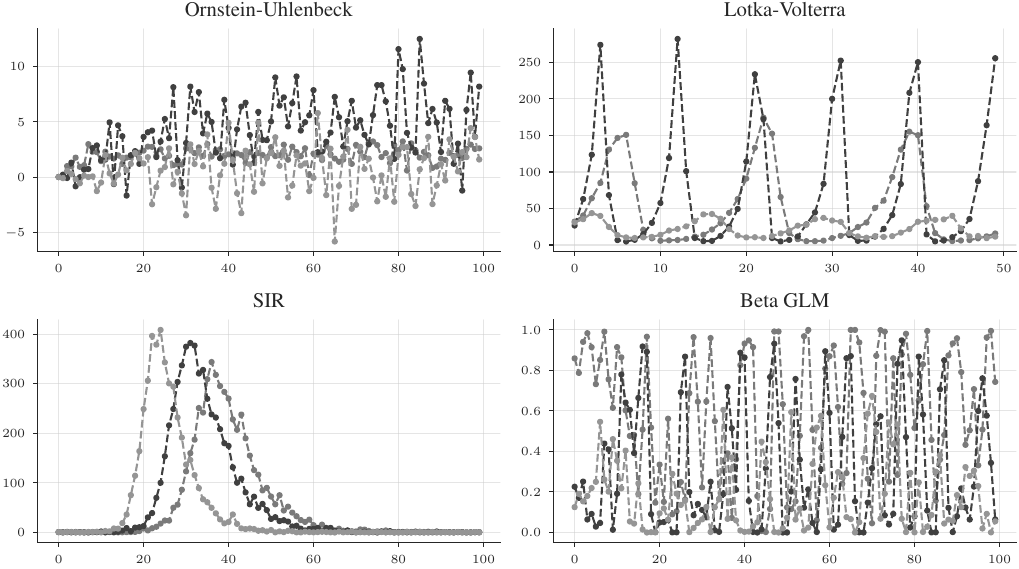}
\caption{Data visualisations of the four benchmark models, Ornstein-Uhlenbeck, Lotka-Volterra, SIR and Beta GLM. For the first three models, the $x$-axis corresponds to the time index of the time series. For the Beta GLM, the $x$-axis only serves as an index.}
\label{fig:sbi_benchmark_data_visualisation}
\end{center}
\end{figure}

\begin{figure}[h!]
\begin{center}
\includegraphics[width=0.65\columnwidth]{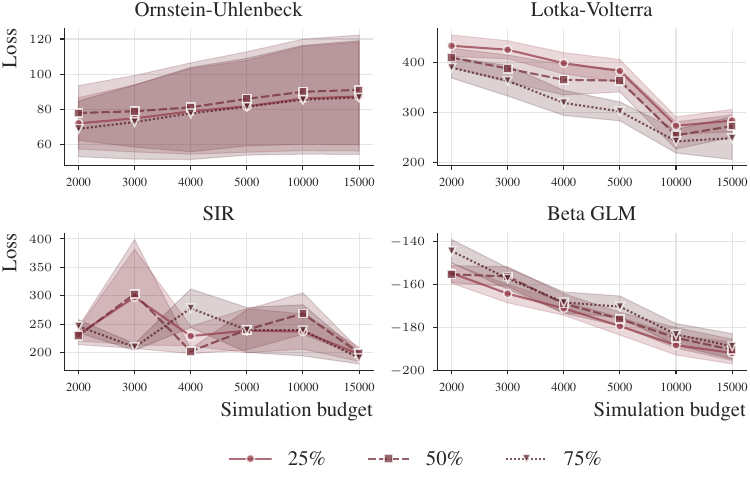}
\caption{Likelihood profiles for the four benchmark models, Ornstein-Uhlenbeck, Lotka-Volterra, SIR and Beta GLM with different reduction factors. Each profile corresponds to the likelihood on the validation set used during training. Using the validation set is a fairly ad-hoc approach and could be done more rigorously, e.g., by using an additional test set where the loss is evaluated instead. Since the data is generated iid, however, this would not arguably not change much. The profiles for these datasets are very similar, only the LV model benefits significantly from different surjection layer dimensionalities.}
\end{center}
\end{figure}

\begin{figure}[h!]
\vskip 0.2in
\begin{center}
\subfloat[H-Min divergences.]{
    \includegraphics[scale=0.675]{fig/sbi_benchmark_divergences.pdf}
}
\subfloat[H-Min divergences (all baselines).]{
    \includegraphics[scale=0.675]{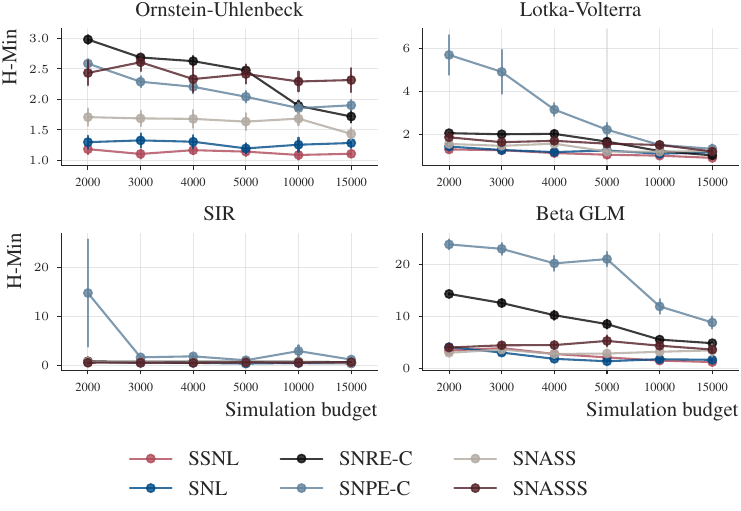}
}
\newline
\subfloat[H-Jensen Shannon divergences.]{    
    \includegraphics[scale=0.675]{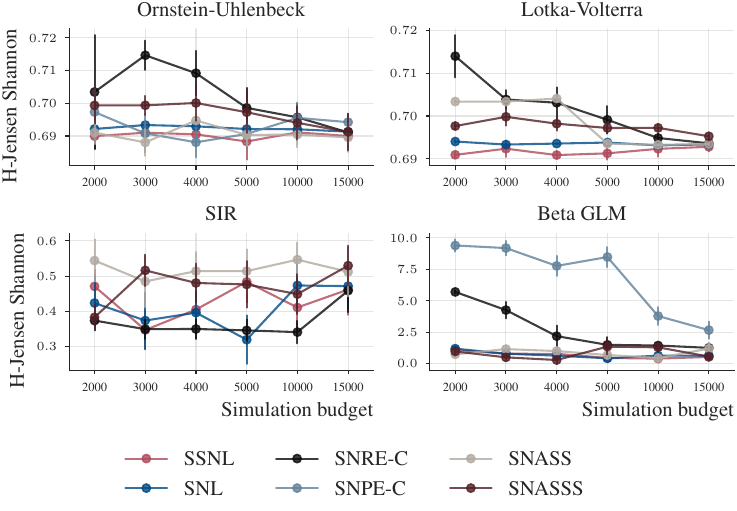}
}
\subfloat[H-Jensen Shannon divergences (all baselines).]{
    \includegraphics[scale=0.675]{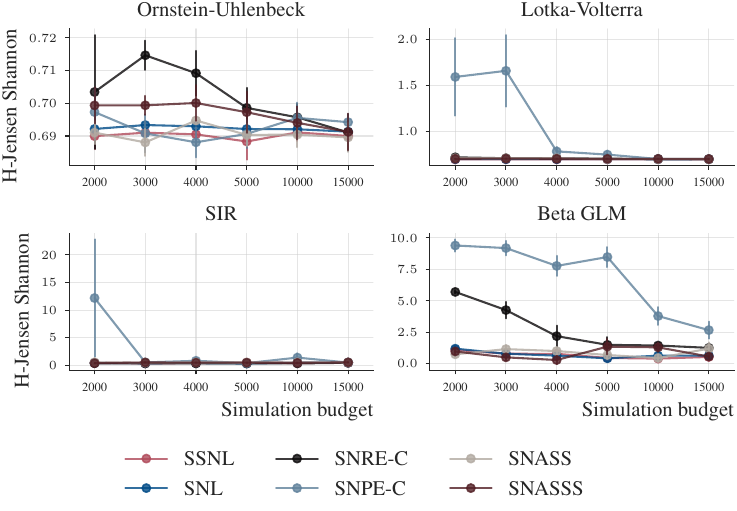}
}
\caption{H-Min and H-Jenson Shannon divergences of the Ornstein-Uhlenbeck, Lotka-Volterra, SIR and Beta GLM models (left withouth SNPE-C for Lotka-Volterra and SIR, right with all baselines). SSNL consistently outperforms all five baselines on Ornstein-Uhlenbeck, Lotka-Volterra, is on par with SNL on Beta GLM and displays mixed results on SIR on both divergence measure. Given that SSNL requires less parameters than SNL, SSNL has the clear advantage in Ornstein-Uhlenbeck, Lotka-Volterra and Beta GLM. The two divergences are consistent for three of the four models, for SIR the H-Min and H-Jensen Shannon show inconsistent divergences.}
\label{app:fig-all-four-benchmarks-results}
\end{center}
\end{figure}

\begin{figure}[h!]

\subsection{Negative examples}
\vskip 0.2in
\begin{center}
\includegraphics[scale=.75]{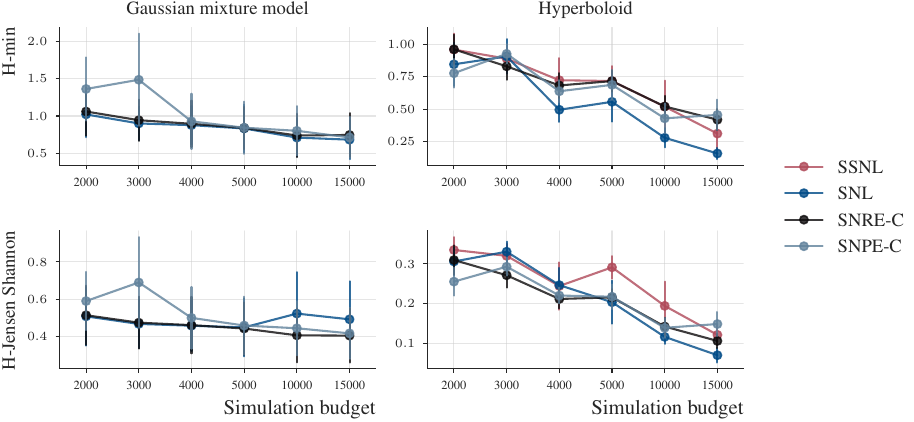}
\caption{Negative examples. We show the H-Min and H-Jensen Shannon divergences on the Gaussian mixture and hyperboloid models. In both cases, SSNL can not outperform the three baselines. Since all data dimensions are informative of posterior parameters, reducing the dimensionality is theoretically only detrimental to the inferences. We did not conduct experiments on SNASS and SNASSS here, since their is no dimensionality reduction necessary.}
\label{fig:negative-examples}
\end{center}
\vskip -0.2in
\end{figure}

\begin{figure}[h!]
\subsection{Solar dynamo model}
\vskip 0.2in
\begin{center}
\includegraphics[scale=.8]{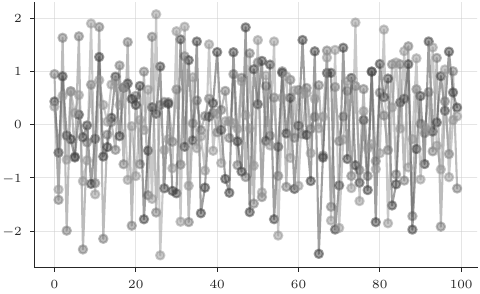}
\caption{Data visualisations of the solar dynamo models. The $x$-axis represents index of the time series $t$, the $y$-axis the observed time point $y_t$.}
\end{center}
\vskip -0.2in
\end{figure}

\begin{figure}[h!]
\begin{center}
\includegraphics[scale=.75]{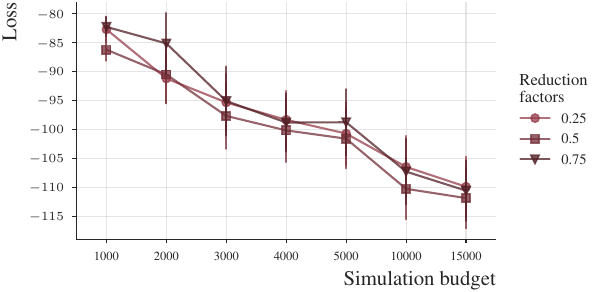}
\caption{Likelihood profiles for the solar dynamo model with different reduction factors. Each profile corresponds to the likelihood on the validation set used during training. The likelihood profiles all show very similar losses (see $y$-axis). As a consequence, we used the model with the greatest reduction on dimensionality, i.e., $25\%$ which reduces the dimensionality in the embedding layer to $Q=25$.}
\end{center}
\end{figure}

\begin{figure}[h!]
\subsection{Jansen-Rit neural mass model}
\vskip 0.2in
\begin{center}
\includegraphics[scale=.8]{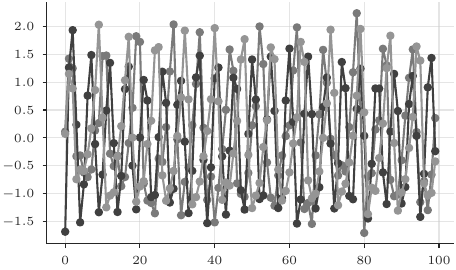}
\caption{Data visualisations of the solar dynamo models. The $x$-axis represents index of the time series $t$, the $y$-axis the observed time point $y_t$.}
\end{center}
\vskip -0.2in
\end{figure}

\begin{figure}[h!]
\begin{center}
\includegraphics[scale=.75]{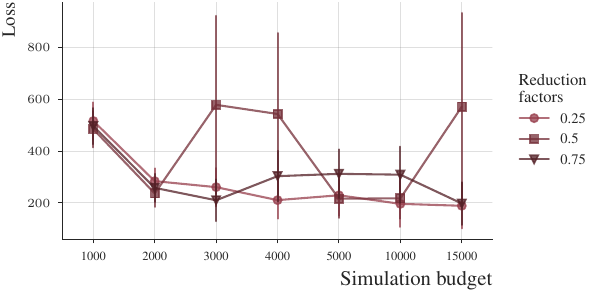}
\caption{Likelihood profiles for the Jansen-Rit model with different reduction factors. Each profile corresponds to the likelihood on the validation set used during training. We used the model with the greatest reduction on dimensionality, i.e., $25\%$, which reduces the dimensionality in the embedding layer to $Q=25$.}
\end{center}
\end{figure}

\end{document}